\documentclass[11pt]{article}
\usepackage[a4paper]{geometry}
\usepackage{clic2017}
\usepackage{times}
\usepackage{url}
\usepackage{latexsym}
\usepackage{todonotes}
\usepackage{graphicx}               

\usepackage{mathtools}
\usepackage{subcaption}
\usepackage{multirow}
\usepackage{arydshln}

\newcommand\blfootnote[1]{%
  \begingroup
  \renewcommand\thefootnote{}\footnote{#1}%
  \addtocounter{footnote}{-1}%
  \endgroup
}

\title{OP-IMS @ DIACR-Ita:\\
Back to the Roots: SGNS+OP+CD still rocks Semantic Change Detection
}

\author{Jens Kaiser, Dominik Schlechtweg, Sabine Schulte im Walde \\
Institute for Natural Language Processing, University of Stuttgart\\
\small
{\tt \{jens.kaiser,schlecdk,schulte\}@ims.uni-stuttgart.de}
}

\date{}

\begin{document}
\maketitle
\begin{abstract}
    We present the results of our participation in the DIACR-Ita shared task on lexical semantic change detection for Italian. We exploit one of the earliest and most influential semantic change detection models based on Skip-Gram with Negative Sampling, Orthogonal Procrustes alignment and Cosine Distance and obtain the winning submission of the shared task with near to perfect accuracy ($.94$). Our results once more indicate that, within the present task setup in lexical semantic change detection, the traditional type-based approaches yield excellent performance.
    \blfootnote{\ \ ``Copyright \textcopyright\ 2020 for this paper by its authors. Use permitted under Creative Commons License Attribution 4.0 International (CC BY 4.0).''}
\end{abstract}

\section{Introduction}
Lexical Semantic Change (LSC) Detection has drawn increasing attention in recent years \cite{kutuzov-etal-2018-diachronic,Tahmasebietal2018Survey}. Recently, SemEval-2020 Task~1 provided a multi-lingual evaluation framework to compare the variety of proposed model architectures \cite{schlechtweg2020semeval}. The DIACR-Ita shared task extends parts of this framework to Italian by providing an Italian data set for SemEval's binary subtask \cite{diacrita_evalita2020}. 

We present the results of our participation in the DIACR-Ita shared task exploiting one of the earliest and most established semantic change detection models based on Skip-Gram with Negative Sampling, Orthogonal Procrustes alignment and Cosine Distance \cite{Hamilton2016a}. Based on our previous research \cite{Schlechtwegetal19,kaiser-etal-2020-IMS} we optimize the dimensionality parameter assuming that high dimensionalities reduce alignment error. With our setting win the shared task with near to perfect accuracy ($.94$). Our results once more demonstrate that, within the present task setup in lexical semantic change detection, the traditional type-based approaches yield excellent performance.

\section{Related Work}

As evident in \newcite{schlechtweg2020semeval} the field of LSCD is currently dominated by Vector Space Models (VSMs), which can be divided into type-based \cite{Turney:2010} and token-based \cite{Schutze1998} models. Prominent type-based models include low-dimensional embeddings such as the Global Vectors \cite[GloVe]{pennington-etal-2014-glove} the Continuous Bag-of-Words (CBOW), the Continuous Skip-gram as well as a slight modification of the latter, the Skip-gram with Negative Sampling model \cite[SGNS]{Mikolov13a,Mikolov13b}. However, as these models come with the deficiency that they aggregate all senses of a word into a single representation, token-based embeddings have been proposed \cite{peters-etal-2018-deep,devlin-etal-2019-bert}.
According to \newcite{Hu19} these models can ideally capture complex characteristics of word use, and how they vary across linguistic contexts. The results of SemEval-2020 Task~1 \cite{schlechtweg2020semeval}, however, show that contrary to this, the token-based embedding models \cite{beck-2020-diasense,kutuzov-giulianelli-2020-uiouva} are heavily outperformed by the type-based ones \cite{prazak-etal-2020-uwb,asgari-etal-2020-emblexchange}. The SGNS model was not only widely used, but also performed best among the participants in the task. Its fast implementation and combination possibilities with different alignment types further solidify SGNS as the standard in LSCD. A common and surprisingly robust \cite{Schlechtwegetal19,kaiser-etal-2020-IMS} practice is to align the time-specific SGNS embeddings with Orthogonal Procrustes (OP) and measure change with Cosine Distance (CD) \cite{Kulkarni14,Hamilton2016b}. This has been shown in several small but independent experiments \cite{Hamilton2016b,Schlechtwegetal19,kaiser-etal-2020-IMS,Shoemark2019} and SGNS+OP+CD has produced two of three top-performing submissions in Subtask 2 in SemEval-2020 Task~1 including the winning submission \cite{pomsl-lyapin-2020-circe,arefyev-zhikov-2020-cmce}.

\section{System overview}

Most VSMs in LSC detection combine three sub-systems: (i) creating semantic word representations, (ii) aligning them across corpora, and (iii) measuring differences between the aligned representations \cite{Schlechtwegetal19}. Alignment is needed as columns from different vector spaces may not correspond to the same coordinate axes, due to the stochastic nature of many low-dimensional word representations \cite{Hamilton2016b}. Following the above-described success, we use SGNS to create word representations in combination with Orthogonal Procrustes (OP) for vector space alignment and Cosine Distance (CD) \cite{SaltonMcGill1983} to measure differences between word vectors. From the resulting graded change predictions we infer binary change values by comparing the target word distribution to the full distribution of change predictions between the target corpora. For our experiments we use the code provided by \newcite{Schlechtwegetal19}.\footnote{\url{https://github.com/Garrafao/LSCDetection}}

\subsection{Semantic Representation}

SGNS is a shallow neural network trained on pairs of word co-occurrences extracted from a corpus with a symmetric window. It represents each word $w$ and each context $c$ as a $d$-dimensional vector to solve

\small
\begin{equation*}
\arg\max_\theta \sum_{(w,c)\in D} \log \sigma(v_c \cdot v_w) + \sum_{(w,c) \in D'} \log \sigma (-v_c \cdot v_w),
\end{equation*}
\normalsize
where $\sigma(x) = \frac{1}{1+e^{-x}}$, $D$ is the set of all observed word-context pairs and $D'$ is the set of randomly generated negative samples \cite{Mikolov13a,Mikolov13b,GoldbergL14}. The optimized parameters $\theta$ are $v_{w_i}$ and $v_{c_i}$ for $i\in 1,...,d$. $D'$ is obtained by drawing $k$ contexts from the empirical unigram distribution $P(c) = \frac{\#(c)}{|D|}$ for each observation of $(w,c)$, cf. \newcite{Levy2015}. After training, each word $w$ is represented by its word vector $v_{w}$. 

Previous research on the influence of parameter settings on SGNS+OP+CD lays the foundation for our parameter choices \cite{Schlechtwegetal19,kaiser-etal-2020-IMS}. Although this sub-system combination is extremely stable regardless of parameter settings, subtle improvements can be achieved by modifying the window size and dimensionality. A common hurdle in LSC detection is the small corpus size, increasing the standard setting for window size from 5 to 10 leads to the creation of more word-context pairs used for training the model. In addition, we also experiment with dimensionalities of $300$ and $500$. Higher dimensionalities alleviate the introduction of noise during the alignment process \cite{kaiser-etal-2020-IMS}. We keep the rest of the parameter settings at their default values (learning rate $\alpha$=$0.025$, \#negative samples $k$=$5$ and sub-sampling $t$=$0.001$).

\subsection{Alignment}
\label{sec:alignment}

SGNS is trained on each corpus separately, resulting in matrices $A$ and $B$. To align them we follow \newcite{Hamilton2016b} and calculate an orthogonally-constrained matrix $W^*$: 
\begin{equation*}
W^* =\underset{W \in O(d)}{\arg\min} \left\lVert B W - A\right\lVert_F
\end{equation*}
where the $i$-th row in matrices $A$ and $B$ correspond to the same word. Using $W^*$ we get the aligned matrices $A^{OP} = A$ and $B^{OP} = BW^*$. Prior to this alignment step we length-normalize and mean-center both matrices \cite{Artetxe2017,Schlechtwegetal19}.

\subsection{Threshold}
\label{sec:threshold}
The DIACR-Ita shared task requires a binary label for each of the target words. However, CD produces graded values between $0.0$ and $2.0$ when measuring differences in word vectors between the two time periods. We tackle this problem by defining a threshold parameter, similar to many approaches applied in SemEval-2020 Task~1 \cite{schlechtweg2020semeval}. All words with a CD greater or equal than the threshold are labeled `1', indicating change. Words with a CD less than the threshold are assigned `0', indicating no change.

A simplified approach is to set the threshold such that the number of words is equal in both groups. This has many disadvantages: Mainly, it relies on the assumption that the two groups are of equal size. This is rarely given in real world applications, especially if the focus is in one word at a time. Thus a more sophisticated approach is needed. In SemEval-2020's Subtask~1 many participants faced the same problem and developed various methods to solve it. Similar to the simplified approach, \newcite{zhou-etal-2020-temporalteller} only look at target words, and after fitting the histogram of CDs to a gamma distribution, set the threshold at the 75\% density quantile. This approach resulted in good performance but is not always applicable due to its dependence on underlying properties of the test set. \newcite{amar-liebeskind-2020-jct} avoid the dependence on target words by randomly selecting 200 words and setting the threshold such that 90\% of the 200 words have a lower distance than the threshold. A more careful selection of words is taken by \newcite{martinc-etal-2020-context}, they look at the CD of semantically stable stop words, accumulate them in different bins and set the threshold to the upper limit of the bin containing fewer than $\#stopwords/\#bins$ words. \newcite{prazak-etal-2020-uwb} propose several methods. One of them is setting the threshold at the mean of the distances of all words in the corpus vocabulary. Our method for determining a threshold is very similar to \newcite{prazak-etal-2020-uwb}, but instead of taking the mean, we use the mean + one standard deviation ($\mu+\sigma$) of all words in the corpus vocabulary.

\section{Experimental setup}
The DIACR-Ita task definition is taken from SemEval-2020 Task~1 Subtask~1 (binary change detection): Given a list of target words and a diacronic corpus pair C$_1$ and C$_2$, the task is to identify the respective target words which have changed their meaning between the time periods t$_1$ and t$_2$ \cite{diacrita_evalita2020,schlechtweg2020semeval}.\footnote{The time periods t$_1$ and t$_2$ were not disclosed to participants.} C$_1$ and C$_2$ have been extracted from Italian newspapers and books. Target words which have changed their meaning are labeled with the value `1', the remaining target words are labeled with `0'. Gold data for the 18 target words is semi-automatically generated from Italian online dictionaries. According to the gold data, 6 of the 18 target words are subject to semantic change between t$_1$ and t$_2$. This gold data was only made public after the evaluation phase. During the evaluation phase each team was allowed to submit 4 predictions for the full list of target words, which were scored using classification accuracy between the predicted labels and the gold data. The final competition ranking compares only the highest of the 4 scores achieved by each team.

\section{Results}

\begin{table}[t]
\centering
\begin{tabular}{cccrrr}
\textbf{entry} & \textbf{dim} & \multicolumn{2}{c}{\textbf{threshold}} & \textbf{ACC} & \textbf{AP} \\
\hline
\#2   & 300  & ($\mu$+$\sigma$) & .76 & \textbf{.944} & .915 \\
\#4   & 500  & ($\mu$+$\sigma$) & .78 &         .889  & .915 \\
\#1   & 300  & (50:50)          & .57 &         .833  & .915  \\
\#3   & 500  & (50:50)          & .64 &         .833  & .915 \\
\cdashline{1-5}
\multicolumn{2}{c}{major. baseline} & -    & & .667 & .333 \\
\multicolumn{2}{c}{freq. baseline} & unk. & & .611 & .418 \\
\multicolumn{2}{c}{colloc. baseline} & unk. && .500 & unk. \\
\end{tabular}
\caption{Accuracy (ACC) and Average Precision (AP) for various parameter settings and thresholds and baselines; \textit{freq. baseline}: Absolute frequency difference between the words in C$_1$ and C$_2$ and an unknown threshold; \textit{colloc. baseline}: Bag of Words + CD and an unknown threshold; \textit{major. baseline}: Every word labeled with `0'.}
\label{tab:results}
\end{table}

\begin{figure*}[ht]
    \begin{subfigure}{0.49\textwidth}
        \includegraphics[width=\linewidth]{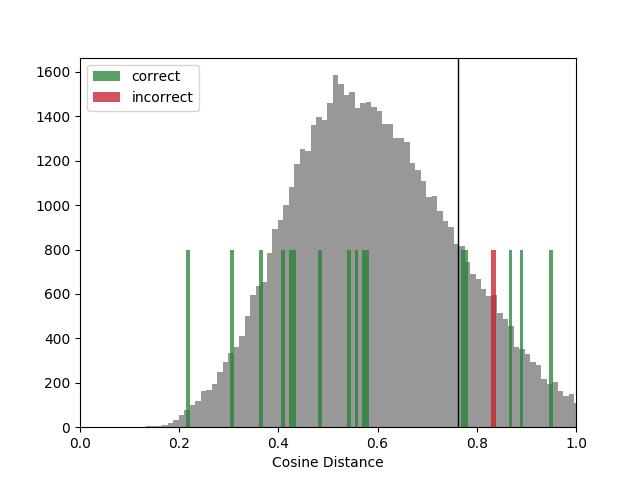}
        \caption{$d$=300}
        \label{fig:hist300}
    \end{subfigure}
    \begin{subfigure}{0.49\textwidth}
        \includegraphics[width=\linewidth]{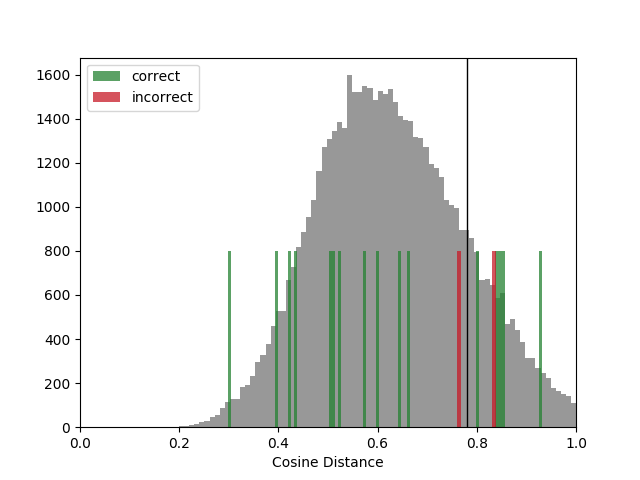}
        \caption{$d$=500}
        \label{fig:hist500}
    \end{subfigure}
    \caption{Background shows histogram (in gray) of CDs for all words in the corpus vocabulary. The colored bars show the CDs of target words, green indicates that the target word was correctly labeled, red indicates incorrect labeling. Vertical line marks threshold value (mean + standard deviation).}
    \label{fig:histogram}
\end{figure*}

We created target word rankings using SGNS+OP+CD with a dimensionality of $300$ and $500$ as described above. From these rankings our predictions are calculated using two different thresholding methods: (i) Splitting the targets into two equally-sized groups (50:50) and (ii) using the mean + one standard deviation ($\mu$+$\sigma$) as threshold, refer to Section \ref{sec:threshold}. The accuracy scores achieved in this way are listed in Table \ref{tab:results}, alongside the official baselines \textit{freq.} and \textit{colloc.} and an additional \textit{major.} baseline. Submission \#2 is our highest scoring submission and won the DIACR-Ita task together with one other undisclosed submission. For both of our rankings the 50:50 threshold yielded lower accuracy than the $\mu$+$\sigma$ threshold. This is due to the imbalance of changed to unchanged target words in the test set. Using $\mu$+$\sigma$ as threshold resulted in an optimal split for the ranking created with $d$=$300$. For $d$=$500$ this threshold was slightly too high with a value of $0.78$. The target word \textit{palmare} which, according to the gold data, has undergone semantic change (label `1') has CD of $0.76$ and was thus incorrectly labeled by our system. Figure \ref{fig:histogram} shows the histogram of CD values for all words of the corpus dictionary in gray. The green and red colored bars correspond target words. If the target word was correctly labeled the bar is green, incorrect labeled target words have red bars. From this visualisation we can see that there is a pronounced gap between the CDs of target words which have changed and those which have not. Our proposed threshold method of $\mu$+$\sigma$ tends to slightly overshoot this gap. This has lead to the lower accuracy of submission \#4, despite the ranking allowing for a higher accuracy. In order to measure the quality of the rankings independent from the threshold we also report AP \cite{Shwartyetal2017} in Table 1, confirming the potential equal performance. 

The method of using the mean + one standard deviation of the CDs of all words in the corpus dictionary resulted in good accuracy, but leaves room for improvement. It tends to over-shoot the gap between unchanged and changed words slightly. Only using the mean shifts the tendency towards under-shooting the gap. The optimal threshold seems to lie somewhere in between. Though, this needs the be confirmed on other, larger, data sets. Furthermore, not all binary classification tasks are suitable for the approach of first creating a ranked list of graded change predictions and then choosing a threshold. The data set of SemEval-2020 Task~1 comprises two tasks, a binary and a ranked task for the same target words. It is not possible to achieve an accuracy of $1$ on the binary task even if all the ranks are predicted correctly for the graded task, i.e., binary change is not just high graded change \cite{schlechtweg2020semeval}.

The one target word which our model labels incorrectly, across a variety of parameter settings, is \textit{piovra}. According to the gold data this word has not undergone semantic change between t$_1$ and t$_2$, while our system labels it as changed. A possible explanation for the error may be  differences in frequency: In C$_1$ \textit{piovra} appears 35 times and in C$_2$ it appears 643 times. SGNS often struggles to create reliable embeddings for low frequency words \cite{kaiser-etal-2020-IMS}. Alternatively, the error could be caused by discrepancies between gold labels and corpora. \newcite{diacrita_evalita2020} state that the gold data is initially based on Italian online dictionaries such as `Sabatini Coletti'. In a manual annotation process the gold data is further refined by providing human judges with up to 100 occurrences of each target word, for which they have to identify the used meaning according to the meanings listed in the dictionaries. A target word is labeled as changed if a meaning is observed in C$_2$ which has not been observed in C$_1$. Although not very likely, it is possible that this annotation method fails to detect novel senses in C$_2$. Sabatini Coletti reports that in addition to the sense ``squid'' \textit{piovra} acquired a new sense ``a secret criminal organisation deeply rooted in society'' in 1983. This might explain why we detect \textit{piovra} as a word which has undergone semantic change given that C$_1$ comprises texts from 1948 to 1970 and C$_2$ comprises texts from 1990 to 2014 \cite{diacrita_evalita2020}.

The DIACR-Ita task dataset is a very valuable contribution to the research field of LSC detection and extends the variety of available data sets to the Italian language. Nonetheless, two points are important when interpreting or results this data set: (i) it contains a small number of target words in combination with binary classification. This makes the data set vulnerable to randomness. (ii) The nature of the gold labels, in addition to possibly not being directly related to the corpus, it is unclear if they reflect semantic change as sense gain and sense loss as in SemEval's Subtask~1. The online dictionaries which create the basis for the gold data only state sense gains. Thus, it might possible for a word to completely lose a sense but still be labeled as unchanged.

\section{Conclusion}
We participated in the DIACR-Ita shared task using well-established type-based methods for diacronic semantic representations in combination with a carefully calculated threshold. We were able to reach the first place with a nearly perfect accuracy of $.94$ confirming once more the reliability of the type-based embeddings created by SGNS, OP as an alignment method and CD to measure differences between word vectors. The presented approach is very suitable for similar tasks as no fine-tuning of parameters is needed. Yet, the system relies on the assumption that graded change is indicative of binary classes.

\section*{Acknowledgments}
Dominik Schlechtweg was supported by the Konrad Adenauer Foundation and the CRETA center funded by the German Ministry for Education and Research (BMBF) during the conduct of this study. We thank the task organizers and reviewers for their efforts.

\bibliographystyle{acl}
\bibliography{biblio}

\end{document}